\pdfoutput=1
\documentclass[11pt]{article}

\usepackage[utf8]{inputenc}
\usepackage[T1]{fontenc}
\usepackage{amsmath,amssymb}
\usepackage{graphicx}
\usepackage{booktabs}
\usepackage{natbib}
\usepackage{hyperref}
\usepackage{xcolor}
\usepackage{array}

\usepackage[margin=1in]{geometry}

\hypersetup{
 colorlinks=true,
 linkcolor=blue!60!black,
 citecolor=blue!60!black,
 urlcolor=blue!60!black,
 pdfauthor={Fatih Uenal, University of Colorado Boulder},
 pdftitle={Swiss-Bench SBP-002: A Frontier Model Comparison on Swiss Legal and Regulatory Tasks},
}

\title{Swiss-Bench SBP-002: A Frontier Model Comparison\\on Swiss Legal and Regulatory Tasks}

\author{
 Fatih Uenal \\
 University of Colorado Boulder \\
 \texttt{fatih.uenal@colorado.edu}
}

\date{}

\begin{document}
\maketitle

\begin{abstract}
While recent work has benchmarked large language models on Swiss legal translation \citep{niklaus2025} and academic legal reasoning from university exams \citep{fan2025}, no existing benchmark evaluates frontier model performance on applied Swiss regulatory compliance tasks. I introduce Swiss-Bench SBP-002, a trilingual benchmark of 395 expert-crafted items spanning three Swiss regulatory domains (FINMA, Legal-CH, EFK), seven task types, and three languages (German, French, Italian), and evaluate ten frontier models from March~2026 using a structured three-dimension scoring framework assessed via a blind three-judge LLM panel (GPT-4o, Claude Sonnet~4, Qwen3-235B) with majority-vote aggregation and weighted $\kappa = 0.605$, with reference answers validated by an independent human legal expert on a 100-item subset (73\% rated Correct, 0\% Incorrect, perfect Legal Accuracy). Results reveal three descriptive performance clusters: Tier~A (35--38\% correct), Tier~B (26--29\%), and Tier~C (13--21\%). The benchmark proves difficult: even the top-ranked model (Qwen~3.5~Plus) achieves only 38.2\% correct, with 47.3\% incorrect and 14.4\% partially correct. Task type difficulty varies widely: legal translation and case analysis yield 69--72\% correct rates, while regulatory Q\&A, hallucination detection, and gap analysis remain below 9\%. Within this roster (seven open-weight, three closed-source), an open-weight model leads the ranking, and several open-weight models match or outperform their closed-source counterparts. These findings provide an initial empirical reference point for assessing frontier model capability on Swiss regulatory tasks under zero-retrieval conditions.
\end{abstract}

\section{Introduction}

The deployment of large language models (LLMs) in legal practice is expanding across jurisdictions \citep{katz2024, rosa2024}. Law firms, compliance departments, and regulatory bodies now use frontier AI systems for legal research, contract analysis, and regulatory interpretation. This raises an empirical question: how reliably can these models perform on jurisdiction-specific legal tasks, particularly in legal systems that differ from the Anglo-American common law tradition on which most training data and evaluation benchmarks are based?

Switzerland presents a hard test case. Its legal system is grounded in the civil law tradition, relying on codified statutes and deductive reasoning rather than case-based precedent. The country operates as a multilingual federation where legislation exists in German, French, Italian, and Romansh, with each version carrying equal legal authority. Swiss regulatory frameworks, spanning the revised Federal Act on Data Protection (nDSG; SR~235.1), financial market supervision under FINMA \citep{finma2024}, and the EU AI Act's impact on Swiss businesses \citep{bundesrat2023}, demand domain knowledge that goes well beyond general legal competence.

Despite recent progress in legal AI evaluation, the benchmarking literature remains oriented toward US legal contexts. LegalBench \citep{guha2023} provides 162 tasks across six reasoning types but is designed around American legal doctrine. LexGLUE \citep{chalkidis2022} standardizes evaluation for legal language understanding but draws primarily from English-language, common law materials. Recent work has begun addressing Swiss legal NLP: SwiLTra-Bench \citep{niklaus2025} benchmarks legal translation across Swiss language pairs, and LEXam \citep{fan2025} evaluates frontier models on 4,886 Swiss law exam questions spanning multiple legal domains. Both, however, target academic legal reasoning (translation fidelity and exam performance, respectively) rather than the applied regulatory compliance scenarios that enterprises face in practice. To the author's knowledge, no existing benchmark evaluates frontier models on the practitioner-oriented tasks central to Swiss regulatory work: interpreting the nDSG for a specific data processing scenario, assessing FINMA circular compliance, or advising on EU AI Act implications for Swiss organizations.

This paper makes three contributions. First, I introduce Swiss-Bench SBP-002\footnote{SBP = Swiss-Bench Paper. SBP-001 was an unpublished 30-item pilot study that informed the dataset design, scoring framework, and judge selection for this work.}, a trilingual benchmark of 395 items across three Swiss regulatory domains and seven task types, designed to evaluate applied regulatory compliance competence, complementing LEXam's academic exam format and SwiLTra-Bench's translation focus with practitioner-oriented compliance scenarios. Second, I develop a structured scoring framework where judges output numeric dimension scores (Legal Accuracy, Citation Accuracy, Completeness) and the grade is computed deterministically, achieving weighted $\kappa = 0.605$ across a three-judge panel, an improvement over traditional grade-mode evaluation. Third, I present a systematic comparison of ten March~2026 frontier models, revealing three descriptive performance clusters with wide task-type difficulty variation.

Section~\ref{sec:related} reviews related work, Section~\ref{sec:methodology} details the benchmark design, Sections~\ref{sec:results}--\ref{sec:discussion} present and interpret results, and Section~\ref{sec:limitations} discusses limitations.

\section{Related Work}
\label{sec:related}

\subsection{LLM Benchmarks for Legal Tasks}

The evaluation of LLMs on legal reasoning has progressed in recent years. LegalBench \citep{guha2023} introduced 162 collaboratively constructed tasks spanning six types of legal reasoning, evaluated across 20 models. The benchmark showed that GPT-4 leads but performance varies widely by reasoning type, with rule-recall and interpretation posing the greatest challenges. LexGLUE \citep{chalkidis2022} provides a standardized suite for legal language understanding but is limited to English-language materials. FinBen \citep{xie2024} provided evaluation across 36 datasets spanning 24 financial tasks at NeurIPS~2024. HELM \citep{liang2023} offered holistic evaluation methodology applicable across domains.

Most relevant here, LEXam \citep{fan2025} recently evaluated frontier LLMs on 4,886 questions from 340 law exams across multiple jurisdictions, including Swiss law. LEXam showed large performance variation across legal domains and jurisdictions, with models achieving near-passing scores on some exams but struggling with jurisdiction-specific nuance. While LEXam provides broad coverage of academic legal reasoning, its exam-based format tests knowledge recall and doctrinal analysis rather than the applied regulatory advisory competence that practitioners require. Swiss-Bench SBP-002 complements LEXam by targeting a different capability: generating actionable compliance guidance for specific regulatory scenarios, scored on structured dimensions (Legal Accuracy, Citation Accuracy, Completeness) that capture regulatory precision rather than exam-style correctness.

\subsection{Swiss Legal NLP}

Research on Swiss legal natural language processing has produced several foundational resources. \citet{niklaus2021} introduced Swiss-Judgment-Prediction, a multilingual dataset for predicting Federal Supreme Court outcomes across German, French, and Italian. LEXTREME \citep{niklaus2023} established a multi-lingual, multi-task benchmark for the legal domain with European coverage, while MultiLegalPile \citep{niklaus2023a} contributed a 689GB multilingual legal corpus. \citet{tyss2024} developed a multilingual dataset for judicial summarization in Switzerland. Most recently, SwiLTra-Bench \citep{niklaus2025} introduced a benchmark of over 180,000 Swiss legal translation pairs across German, French, Italian, and Romansh, establishing the first systematic evaluation of LLM translation quality for Swiss legal texts.

Together with LEXam (Section~\ref{sec:related}.1), these contributions cover legal knowledge retrieval, judgment prediction, translation, and summarization. None of these, however, evaluates the applied regulatory advisory capability that compliance professionals require: synthesizing domain knowledge into actionable guidance for specific Swiss regulatory scenarios. This paper addresses that gap by benchmarking frontier models on practitioner-oriented tasks across three Swiss regulatory domains, using a structured scoring framework with deterministic grading.

\subsection{LLM-as-Judge Methodology}

LLMs are now widely used as automated evaluators, serving as a scalable alternative to human expert assessment. \citet{zheng2023} introduced the MT-Bench and Chatbot Arena frameworks, showing that strong LLM judges can approximate human preferences with over 80\% agreement, while identifying systematic biases (position bias, verbosity bias, self-enhancement bias) that must be addressed through careful protocol design. Arena-Hard-Auto \citep{li2024arena} extended this to multi-judge ensembles for automated benchmarking. G-Eval \citep{liu2023geval} showed that rubric-based LLM evaluation with structured scoring achieves stronger human alignment than unstructured assessment. I adopt the multi-judge ensemble approach of \citet{zheng2023} and mitigates known biases through three mechanisms: (1)~a heterogeneous judge panel drawn from three distinct model families (OpenAI, Anthropic, Alibaba), reducing systematic bias from any single provider; (2)~blind evaluation with anonymized responses; and (3)~structured numeric scoring with deterministic grade computation, which removes the subjective grading boundary from judges and improves inter-rater reliability (weighted $\kappa = 0.605$; see Section~\ref{sec:judges}).

\subsection{AI Compliance and Regulatory Frameworks}

The EU AI Act has spurred work on evaluating LLM compliance capabilities. The COMPL-AI Framework \citep{guldimann2024} provides a technical interpretation and benchmarking suite aligned with EU AI Act requirements. \citet{catillo2024} explored assurance of EU AI Act compliance and adversarial robustness. \citet{stern2024} addressed multilingual legal reasoning for judicial support across European jurisdictions. The Apertus initiative \citep{initiative2025} seeks to democratize open and compliant LLMs for diverse language environments, including Swiss languages. These developments motivate the inclusion of EU AI Act impact assessment as a dedicated domain in Swiss-Bench SBP-002, reflecting the practical reality that Swiss organizations must navigate both domestic and European regulatory requirements.

\section{Methodology}
\label{sec:methodology}

This section describes the dataset construction (\S\ref{sec:dataset}), model selection (\S\ref{sec:models}), structured scoring framework (\S\ref{sec:scoring}), multi-judge evaluation panel (\S\ref{sec:judges}), and statistical analysis (\S\ref{sec:stats}).

\subsection{Dataset Construction and Composition}
\label{sec:dataset}

The evaluation dataset comprises 395 items across three Swiss regulatory domains and three languages, selected via stratified sampling (seed=42) from a larger pool of approximately 2,000 candidate items, with target quotas across domains, difficulty levels, and task types, though the realized composition is not fully balanced (see Table~\ref{tab:crosstab} for the actual distribution). The construction follows a multi-phase verification pipeline inspired by the data curation methodology of OpenScholar \citep{asai2026}.

\textbf{Domain coverage.} Items span three regulatory domains reflecting the core areas of Swiss AI compliance practice:
\begin{itemize}
    \item \textbf{FINMA} (178 items): Financial market regulation under FINMA circulars, the Banking Act (BankG; SR~952.0), anti-money laundering provisions, and sector-specific AI deployment requirements.
    \item \textbf{Legal-CH} (169 items): Swiss federal law including the revised Federal Act on Data Protection (nDSG; SR~235.1), the Code of Obligations (OR), and cross-jurisdictional regulatory alignment with EU frameworks including the EU AI Act.
    \item \textbf{EFK (Reg.\ QA)} (48 items): Swiss Federal Audit Office requirements, internal control standards, and audit-specific AI governance scenarios.
\end{itemize}

\textbf{Trilingual design.} Following SwiLTra-Bench \citep{niklaus2025}, which found large performance variation across Swiss language pairs, items are distributed across German (150), French (148), and Italian (97), reflecting Switzerland's three primary official languages. French and Italian items were translated from German source items using GPT-4o. Each translation was subjected to a legal terminology audit: the author verified that (1)~statutory references (article numbers, law abbreviations) were preserved correctly, (2)~domain-specific terms matched the official terminology used in the corresponding language version of Swiss federal legislation, and (3)~no substantive meaning was altered by the translation. Items failing any criterion were re-translated with corrective prompts. This audit was conducted by the author, not by an independent legal translator, which is a limitation acknowledged in Section~\ref{sec:limitations}. An additional 18 English items, translated from German, are included to enable future cross-lingual calibration studies but are excluded from all reported analyses.

Table~\ref{tab:dataset} summarizes the dataset composition by domain and language.

\begin{table}[h]
\centering
\caption{Swiss-Bench v3 dataset composition (395 main items). 18 English experimental items excluded from main analysis.}
\label{tab:dataset}
\begin{tabular}{lcccc}
\toprule
\textbf{Domain} & \textbf{DE} & \textbf{FR} & \textbf{IT} & \textbf{Total} \\
\midrule
FINMA & 80 & 62 & 36 & 178 \\
Legal-CH & 40 & 74 & 55 & 169 \\
EFK (Reg.\ QA) &30 & 12 & 6 & 48 \\
\midrule
\textbf{Total} & \textbf{150} & \textbf{148} & \textbf{97} & \textbf{395} \\
\bottomrule
\end{tabular}
\end{table}

\textbf{Task type taxonomy.} Items are classified into seven task types, each targeting a distinct legal reasoning capability:
\begin{enumerate}
    \item \textit{Regulatory Q\&A} (104 items): Direct questions about regulatory provisions requiring accurate statutory interpretation.
    \item \textit{Hallucination detection} (63 items): Scenarios containing plausible but incorrect legal claims that models must identify and correct.
    \item \textit{Regulatory gap analysis} (59 items): Tasks requiring identification of compliance gaps between current practice and regulatory requirements.
    \item \textit{Jurisdiction discrimination} (58 items): Scenarios testing whether models correctly distinguish Swiss law from structurally similar EU or international frameworks.
    \item \textit{Statutory interpretation} (46 items): Complex interpretive questions requiring synthesis across multiple statutory provisions.
    \item \textit{Case analysis} (35 items): Fact-pattern analysis requiring application of regulatory frameworks to specific scenarios.
    \item \textit{Legal translation} (30 items): Cross-lingual regulatory terminology tasks across DE/FR/IT.
\end{enumerate}

\textbf{Ground truth construction.} Each item includes an expert-authored reference answer constructed through a three-phase pipeline: (1)~initial drafting by the author based on primary statutory sources, (2)~adversarial verification where edge cases and ambiguities are identified and resolved, and (3)~quality-gated filtering where items failing clarity or specificity thresholds are revised or removed. This approach parallels the adversarial multi-phase verification used in OpenScholar's data curation \citep{asai2026}. The single-author design is a limitation acknowledged in Section~\ref{sec:limitations}.

\subsection{Model Selection}
\label{sec:models}

Ten models were evaluated, selected to represent the March~2026 frontier across two categories: \emph{closed-source} (model weights not publicly released; accessed only via proprietary API) and \emph{open-weight} (model weights publicly available for download, regardless of license restrictions) (Table~\ref{tab:models}). Nine models were accessed via OpenRouter; Gemini~2.5~Flash was accessed via the Google AI API directly. All evaluations used temperature~0 and maximum output length of 4,096 tokens. No system prompt or retrieval augmentation was used; each prompt was submitted as a standalone user message to ensure uniform evaluation conditions. The benchmark therefore evaluates parametric legal knowledge, not retrieval-based reasoning.

\begin{table}[h]
\centering
\caption{Models evaluated in SBP-002.}
\label{tab:models}
\begin{tabular}{llll}
\toprule
\textbf{Model} & \textbf{Provider} & \textbf{Category} & \textbf{API Identifier} \\
\midrule
Claude Sonnet 4 & Anthropic & Closed-source & anthropic/claude-sonnet-4 \\
Gemini 2.5 Flash & Google & Closed-source & google/gemini-2.5-flash \\
GPT-4o & OpenAI & Closed-source & openai/gpt-4o \\
GLM 5 & Zhipu AI & Open-weight & z-ai/glm-5 \\
Qwen 3.5 Plus & Alibaba & Open-weight & qwen/qwen3.5-plus-02-15 \\
MiMo-V2-Flash & Xiaomi & Open-weight & xiaomi/mimo-v2-flash \\
GPT-oss 120B & OpenAI & Open-weight & openai/gpt-oss-120b \\
MiniMax M2.5 & MiniMax & Open-weight & minimax/minimax-m2.5 \\
Mistral Large 3 & Mistral & Open-weight & mistralai/mistral-large-2411 \\
DeepSeek V3.2 & DeepSeek & Open-weight & deepseek/deepseek-chat \\
\bottomrule
\end{tabular}
\end{table}

\subsection{Structured Scoring Framework}
\label{sec:scoring}

I adopt a structured numeric scoring approach to improve inter-rater reliability over traditional grade-mode evaluation, consistent with the rubric-driven methodology of ScholarQABench \citep{asai2026}. Rather than asking judges to assign a holistic letter grade, each judge outputs three dimension scores as a JSON object, and the final grade is computed deterministically.

\textbf{Scoring dimensions.} Each model response is evaluated on three dimensions:
\begin{enumerate}
    \item \textbf{Legal Accuracy} (weight 0.5): Correctness of legal claims, valid statutory citations, absence of fabricated provisions.
    \item \textbf{Citation Accuracy} (weight 0.3): Whether cited articles, circulars, and provisions exist and are correctly attributed to the relevant Swiss legal framework.
    \item \textbf{Completeness} (weight 0.2): Coverage of all aspects raised in the prompt, including relevant sub-issues and cross-references.
\end{enumerate}

Each dimension is scored on a three-point scale $\{0.0, 0.5, 1.0\}$. The combined score is computed as:
\begin{equation}
    S = 0.5 \cdot \text{Legal\_Accuracy} + 0.3 \cdot \text{Citation\_Accuracy} + 0.2 \cdot \text{Completeness}
\end{equation}

The final grade is assigned deterministically:
\begin{itemize}
    \item \textbf{C} (Correct): $S \geq 0.8$
    \item \textbf{P} (Partially correct): $0.5 \leq S < 0.8$
    \item \textbf{I} (Incorrect): $S < 0.5$
\end{itemize}

The thresholds (0.8 for C, 0.5 for P) and weights (0.5/0.3/0.2) were chosen to prioritize legal accuracy while giving meaningful weight to citation correctness. Alternative threshold choices (e.g., 0.7/0.4) would shift the absolute C-rates but are unlikely to change relative model rankings, since the ranking is driven by the ordering of combined scores rather than any single threshold. A formal threshold-sensitivity analysis is left to future work. This structured approach removes the subjective partial-credit boundary from judges, who assess only factual dimensions. In calibration testing, the same 30 items were scored by the same three judges under both modes (grade mode first, structured mode second; 30 items $\times$ 3 judges = 90 scored pairs per mode). Structured scoring improved weighted Cohen's $\kappa$ from 0.512 to 0.605, primarily by reducing P$\leftrightarrow$I boundary disputes by 17\% and narrowing the leniency gap between judges. Order effects were not controlled (structured mode always followed grade mode), which is a limitation of this calibration comparison.

\subsection{Multi-Judge Evaluation Panel}
\label{sec:judges}

Following the multi-judge ensemble methodology of MT-Bench \citep{zheng2023} and Arena-Hard-Auto \citep{li2024arena}, evaluation is conducted by a panel of three heterogeneous LLM judges spanning distinct model families:
\begin{enumerate}
    \item \textbf{GPT-4o} (OpenAI)
    \item \textbf{Claude Sonnet 4} (Anthropic)
    \item \textbf{Qwen3-235B} (Alibaba)
\end{enumerate}

Each judge independently scores every response on all three dimensions via structured JSON output. The final grade for each item is determined by majority vote across the three judges' independently computed C/P/I grades. Majority vote on discrete grades was preferred over averaging the continuous weighted scores because it preserves each judge's independent threshold application and is more robust to outlier dimension scores from a single judge. All responses are anonymized prior to evaluation: model names, provider-specific formatting artifacts, and identifying metadata are stripped, and each response receives an opaque identifier.

\textbf{Judge selection.} The upgrade from cost-efficient judges (Haiku~4.5, GPT-4o-mini, Gemini~2.0~Flash in Round~1) to frontier-class judges addresses a limitation identified in the pilot study: cost-efficient judges may share blind spots with evaluated models on highly specialized Swiss regulatory questions. The current panel provides stronger evaluation depth while maintaining cross-provider diversity.

\begin{figure}[h]
\centering
\includegraphics[width=0.85\textwidth]{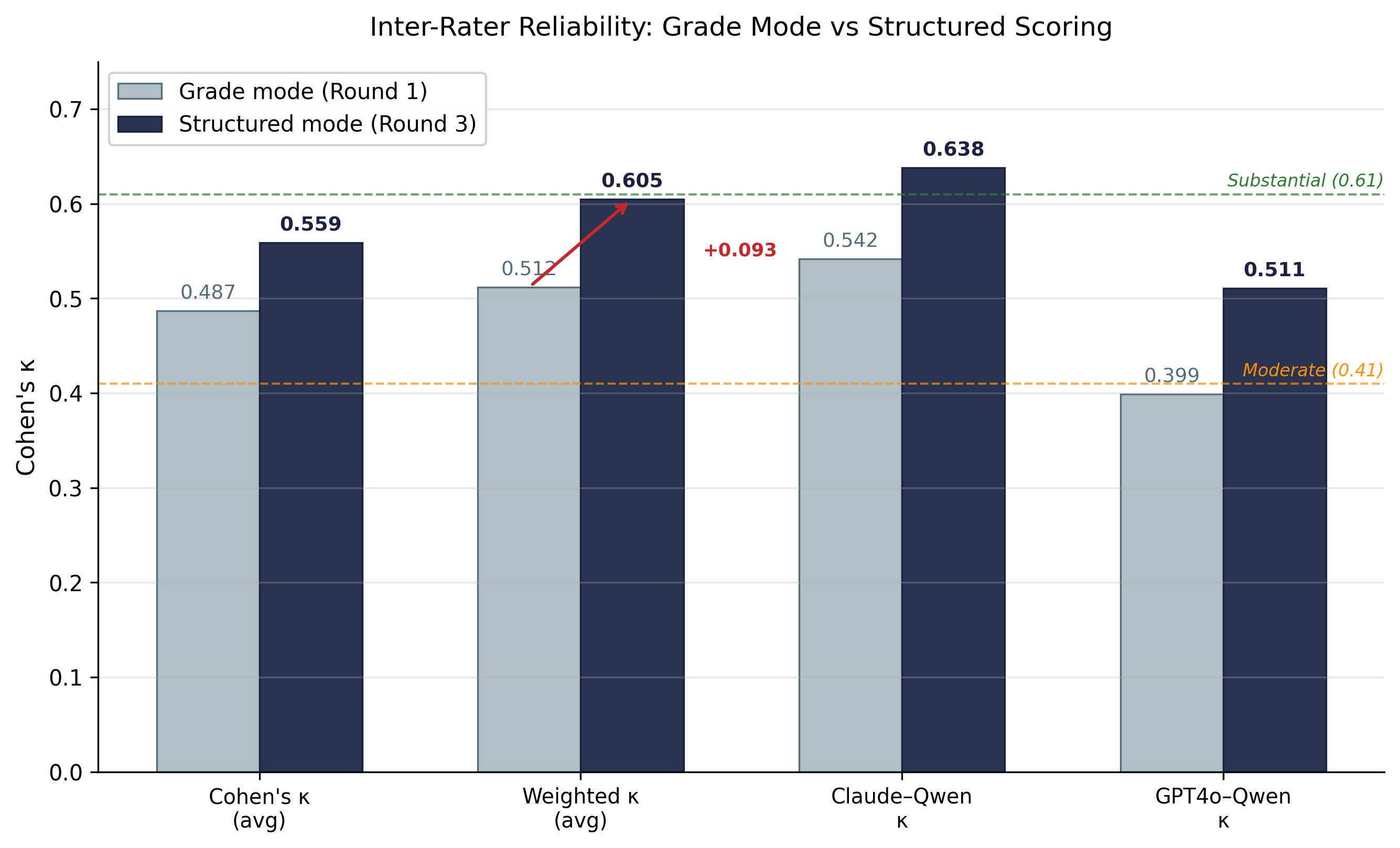}
\caption{Inter-rater reliability improvement from grade-mode to structured scoring. Dashed lines indicate Landis \& Koch interpretation thresholds. Structured scoring moves the panel average from moderate ($\kappa = 0.512$) to substantial ($\kappa = 0.605$) agreement.}
\label{fig:irr}
\end{figure}

\textbf{Inter-rater reliability.} Agreement is assessed using linearly weighted Cohen's $\kappa$ \citep{landis1977} for each judge pair on a 30-item calibration subset (the same items used for the grade-mode vs.\ structured-mode comparison in Section~\ref{sec:scoring}). The pairwise weighted $\kappa$ values are: Claude Sonnet~4 $\times$ Qwen3-235B $\kappa = 0.707$, GPT-4o $\times$ Qwen3-235B $\kappa = 0.556$, and GPT-4o $\times$ Claude Sonnet~4 $\kappa = 0.553$. The panel average (arithmetic mean of pairwise values) is $\kappa = 0.605$ (at the moderate--substantial boundary on the Landis \& Koch scale). A multi-rater statistic such as Fleiss' $\kappa$ was not computed; future work should supplement the pairwise analysis. On the calibration subset, 76\% of items received unanimous agreement across all three judges, and the primary source of remaining disagreement (91\% of cases) is the P$\leftrightarrow$I boundary, consistent with known difficulty in human--human agreement on subjective annotation tasks \citep{artstein2008}. Figure~\ref{fig:irr} illustrates the improvement across all judge pairs.

\subsection{Statistical Analysis}
\label{sec:stats}

The primary metric is the correct rate (C\%): the proportion of items receiving a C grade via majority vote across the three-judge panel. This item-level metric follows the holistic evaluation approach of HELM \citep{liang2023}, applied at the level of individual regulatory scenarios rather than aggregated dimensions. Confidence intervals are computed using the Wilson score method \citep{brown2001}, which provides better coverage than the Wald interval for proportions near 0 or 1. To formally test whether model differences are statistically significant while accounting for item-level difficulty variation, I fit a mixed-effects linear probability model: $\text{correct}_{ij} \sim \text{model}_j + (1 \mid \text{item}_i)$, where the random intercept captures item difficulty. Inter-rater reliability is assessed via linearly weighted Cohen's $\kappa$ \citep{landis1977} for all three judge pairs.

\section{Results}
\label{sec:results}

\subsection{Overall Rankings and Tier Structure}

\begin{figure}[h]
\centering
\includegraphics[width=0.85\textwidth]{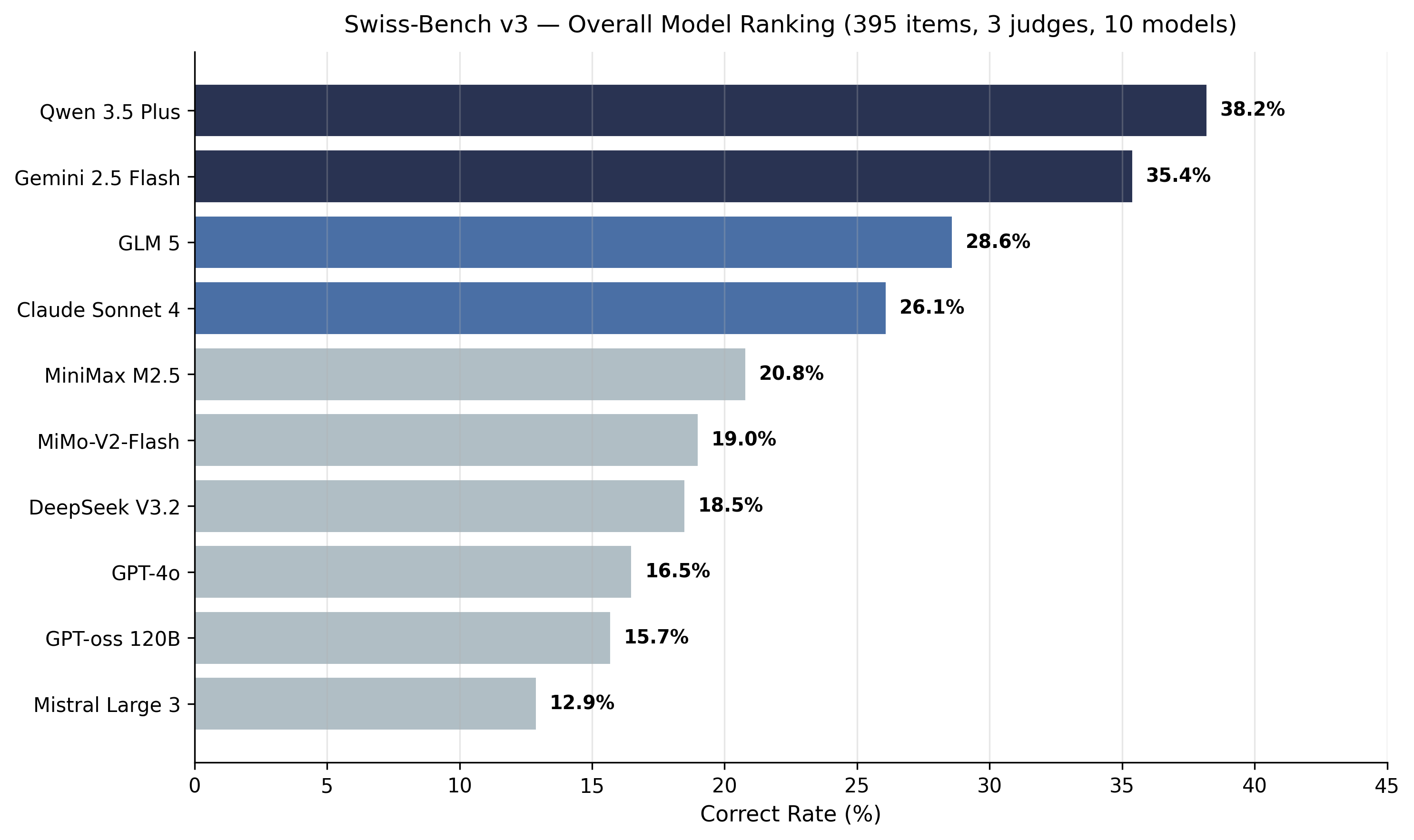}
\caption{Overall model ranking by correct rate on 395 items. Colors indicate tier: dark (A), medium (B), light (C).}
\label{fig:ranking}
\end{figure}

Table~\ref{tab:rankings} presents the overall model rankings by correct rate (C\%). Each of the 395 items receives a C/P/I grade via majority vote across the three-judge panel. Figure~\ref{fig:ranking} visualizes the ranking.

\begin{table}[h]
\centering
\caption{Overall model rankings by correct rate (C\%) on 395 items with 95\% Wilson score confidence intervals. P = partially correct, I = incorrect. Tier assignment based on natural breaks in the C\% distribution.}
\label{tab:rankings}
\begin{tabular}{clcccccc}
\toprule
\textbf{Rank} & \textbf{Model} & \textbf{Category} & \textbf{C\%} & \textbf{95\% CI} & \textbf{P\%} & \textbf{I\%} & \textbf{Tier} \\
\midrule
1 & Qwen 3.5 Plus & Open-weight & 38.2 & [33.6, 43.1] & 14.4 & 47.3 & A \\
2 & Gemini 2.5 Flash & Closed & 35.4 & [30.9, 40.3] & 16.5 & 48.1 & A \\
\midrule
3 & GLM 5 & Open-weight & 28.6 & [24.4, 33.3] & 14.7 & 56.7 & B \\
4 & Claude Sonnet 4 & Closed & 26.1 & [22.0, 30.6] & 14.7 & 59.2 & B \\
\midrule
5 & MiniMax M2.5 & Open-weight & 20.8 & [17.1, 25.0] & 13.7 & 65.6 & C \\
6 & MiMo-V2-Flash & Open-weight & 19.0 & [15.4, 23.1] & 10.1 & 70.9 & C \\
7 & DeepSeek V3.2 & Open-weight & 18.5 & [15.0, 22.6] & 15.2 & 66.3 & C \\
8 & GPT-4o & Closed & 16.5 & [13.1, 20.4] & 18.0 & 65.6 & C \\
9 & GPT-oss 120B & Open-weight & 15.7 & [12.4, 19.6] & 9.6 & 74.7 & C \\
10 & Mistral Large 3 & Open-weight & 12.9 & [10.0, 16.6] & 13.4 & 73.7 & C \\
\bottomrule
\end{tabular}
\end{table}

Three descriptive clusters emerge based on natural breaks in the C\% distribution: Tier~A comprises two models at 35--38\% C-rate, Tier~B two models at 26--29\%, and Tier~C six models at 13--21\%. The largest gap (6.8 percentage points) separates Tier~A from Tier~B, and the second-largest (5.3 points) separates Tier~B from Tier~C. The Wilson confidence intervals are consistent with the cluster assignment: the Tier~A and Tier~B confidence intervals show minimal overlap (Gemini lower bound 30.9\% vs.\ GLM upper bound 33.3\%), which is consistent with the A--B boundary, though this observation does not constitute a formal clustering test. Within tiers, adjacent models' CIs overlap, indicating that fine-grained rank differences (e.g., rank~5 vs.\ rank~6) should not be over-interpreted.

\textbf{Mixed-effects analysis.} A linear probability model with random item intercepts ($\text{correct}_{ij} \sim \text{model}_j + (1 \mid \text{item}_i)$; $N = 3{,}950$ observations, 395 items, 10 models) reveals that item difficulty accounts for 47.2\% of the variance (ICC = 0.472), confirming that what is being asked matters as much as which model answers. All Tier~A and Tier~B models are significantly more accurate than the lowest-ranked model (Mistral Large~3) at $p < 10^{-4}$. In contrast, GPT-4o ($p = 0.098$) and GPT-oss~120B ($p = 0.193$) are not statistically distinguishable from the baseline, indicating that the bottom three models form a statistically indistinguishable cluster. The Tier~A--B boundary is supported: both Tier~A models show significantly larger effects (Qwen: $+0.253$, $p < 10^{-32}$; Gemini: $+0.225$, $p < 10^{-25}$) than Tier~B models (GLM: $+0.157$, $p < 10^{-13}$; Claude: $+0.132$, $p < 10^{-9}$). The overall incorrect rates are high (47--75\%), reflecting the difficulty of the benchmark: the structured scoring framework with domain-expert reference answers sets a higher bar than generic LLM evaluation.

Both Tier~A models are from different provider families (Alibaba, Google), and both outperform all three closed-source Western models (Claude Sonnet~4 rank~4, GPT-4o rank~8). Second, seven of ten models are open-weight, yet the three closed-source models occupy ranks~2, 4, and 8; closed-source status does not confer an advantage on this benchmark. Third, the high I-rates across all models (47\% even for the top-ranked Qwen~3.5~Plus) indicate that no current model produces reliable outputs on isolated Swiss regulatory prompts under zero-retrieval conditions. Performance in realistic workflows with retrieval augmentation, tool use, or domain-specific fine-tuning may differ.

\subsection{Per-Domain Analysis}

\begin{figure}[h]
\centering
\includegraphics[width=0.85\textwidth]{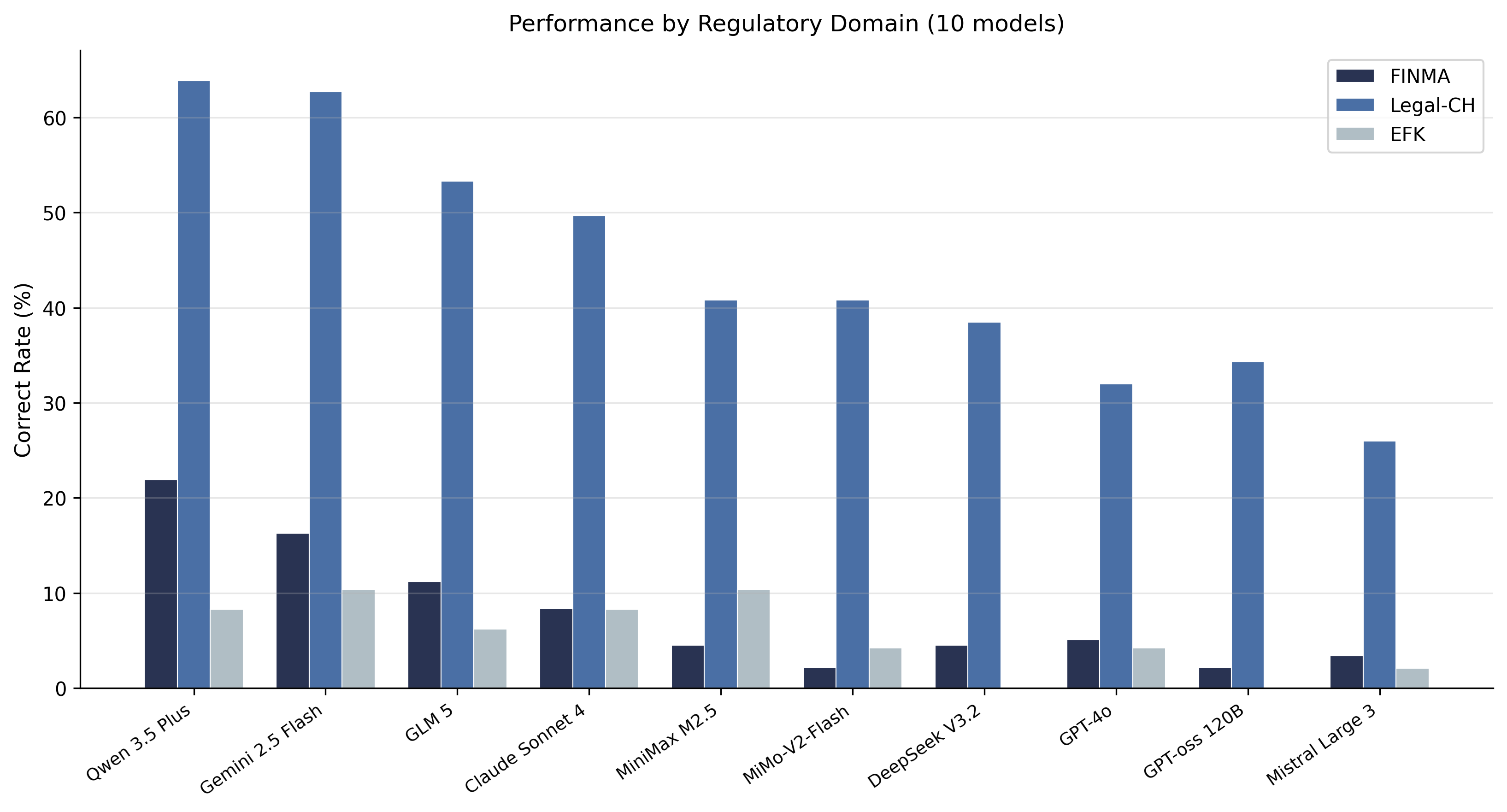}
\caption{Correct rate by regulatory domain. Legal-CH is much easier than FINMA and EFK (Reg.\ QA) across all models. Note: EFK results are confounded with the Regulatory Q\&A task type (all 48 EFK items belong to this single task type).}
\label{fig:domains}
\end{figure}

Performance varies widely across the three regulatory domains (Figure~\ref{fig:domains}).

\begin{table}[h]
\centering
\caption{Correct rate (C\%) by model and regulatory domain.}
\label{tab:domains}
\begin{tabular}{lccc}
\toprule
\textbf{Model} & \textbf{FINMA} & \textbf{Legal-CH} & \textbf{EFK (Reg.\ QA)} \\
\midrule
Qwen 3.5 Plus     & 21.9 & 63.9 & 8.3 \\
Gemini 2.5 Flash   & 16.3 & 62.7 & 10.4 \\
GLM 5              & 11.2 & 53.3 & 6.2 \\
Claude Sonnet 4    & 8.4  & 49.7 & 8.3 \\
MiniMax M2.5       & 4.5  & 40.8 & 10.4 \\
MiMo-V2-Flash      & 2.2  & 40.8 & 4.2 \\
DeepSeek V3.2      & 4.5  & 38.5 & 0.0 \\
GPT-4o             & 5.1  & 32.0 & 4.2 \\
GPT-oss 120B       & 2.2  & 34.3 & 0.0 \\
Mistral Large 3    & 3.4  & 26.0 & 2.1 \\
\midrule
\textbf{Mean}      & \textbf{8.0} & \textbf{44.2} & \textbf{5.4} \\
\bottomrule
\end{tabular}
\end{table}

Legal-CH is the most accessible domain (mean C\% = 44.2\%), while FINMA (8.0\%) and EFK (5.4\%) prove extremely challenging. Even the top-ranked model achieves only 21.9\% on FINMA and 8.3\% on EFK. The Legal-CH advantage likely reflects broader coverage of Swiss federal law (nDSG, OR) in pre-training corpora, while FINMA circulars and EFK audit standards represent specialized regulatory knowledge with limited public documentation. The near-zero EFK performance for DeepSeek~V3.2 and GPT-oss~120B (0.0\% C) suggests these models have effectively no knowledge of Swiss Federal Audit Office requirements. Note that all 48 EFK items belong to the regulatory Q\&A task type; EFK domain performance is therefore confounded with task-type difficulty and should not be interpreted as a general measure of EFK capability.

\subsection{Per-Task-Type Analysis}

\begin{figure}[h]
\centering
\includegraphics[width=0.92\textwidth]{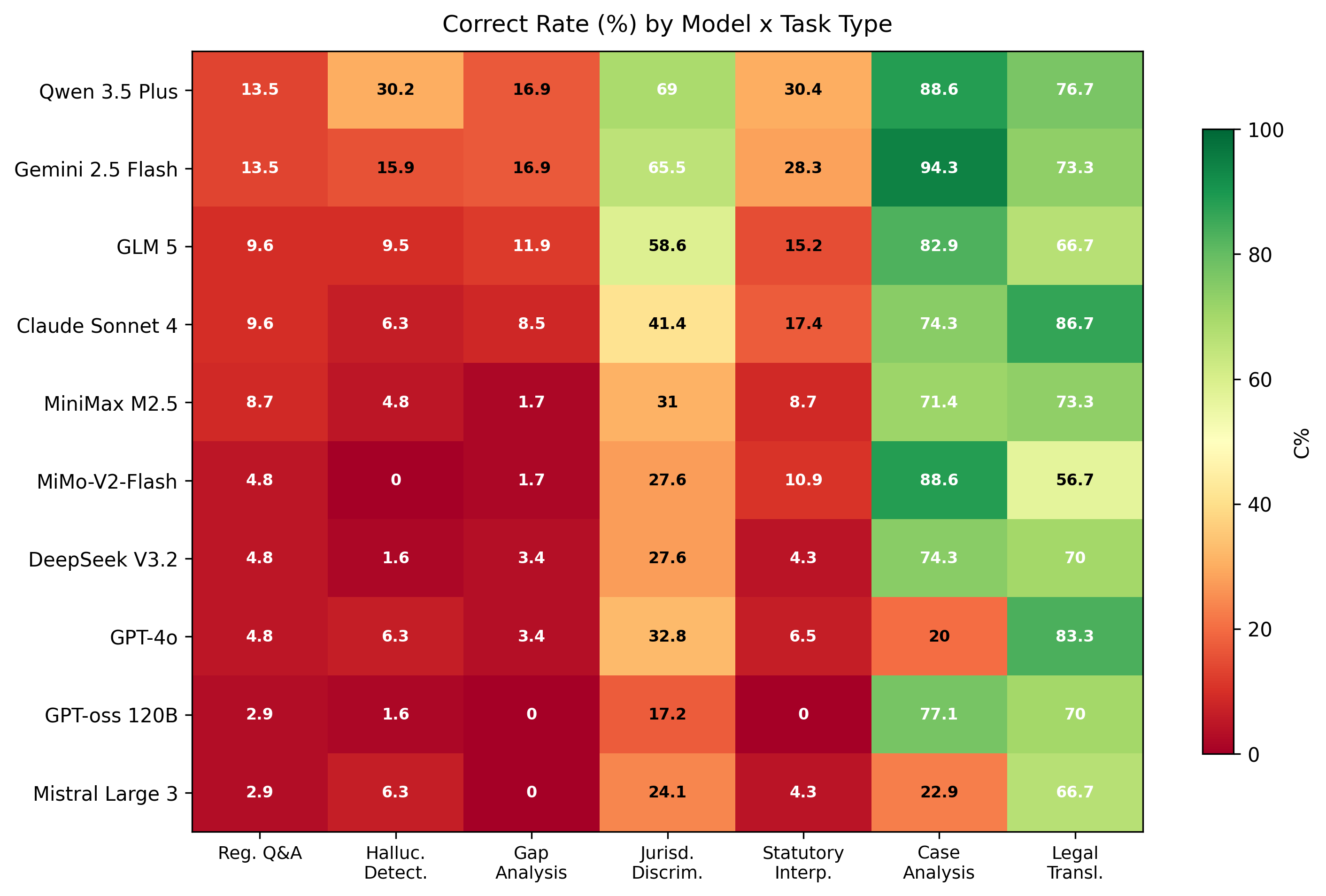}
\caption{Heatmap of correct rate (\%) by model and task type. Case analysis and legal translation are easiest; regulatory Q\&A and gap analysis are hardest.}
\label{fig:heatmap}
\end{figure}

The seven task types reveal a clear difficulty gradient (Figure~\ref{fig:heatmap}). Legal translation (72.3\% mean C) and case analysis (69.4\%) are the easiest, while regulatory Q\&A (7.5\%), hallucination detection (8.3\%), and regulatory gap analysis (6.4\%) are the hardest.

\begin{table}[h]
\centering
\caption{Correct rate (C\%) by task type. Models sorted by overall rank.}
\label{tab:tasks}
\begin{tabular}{lccccccc}
\toprule
\textbf{Model} & \textbf{Reg.} & \textbf{Halluc.} & \textbf{Gap} & \textbf{Jurisd.} & \textbf{Stat.} & \textbf{Case} & \textbf{Legal} \\
 & \textbf{Q\&A} & \textbf{Det.} & \textbf{Anal.} & \textbf{Discr.} & \textbf{Interp.} & \textbf{Anal.} & \textbf{Transl.} \\
\midrule
Qwen 3.5 Plus    & 13.5 & 30.2 & 16.9 & 69.0 & 30.4 & 88.6 & 76.7 \\
Gemini 2.5 Flash  & 13.5 & 15.9 & 16.9 & 65.5 & 28.3 & 94.3 & 73.3 \\
GLM 5             & 9.6  & 9.5  & 11.9 & 58.6 & 15.2 & 82.9 & 66.7 \\
Claude Sonnet 4   & 9.6  & 6.3  & 8.5  & 41.4 & 17.4 & 74.3 & 86.7 \\
MiniMax M2.5      & 8.7  & 4.8  & 1.7  & 31.0 & 8.7  & 71.4 & 73.3 \\
MiMo-V2-Flash     & 4.8  & 0.0  & 1.7  & 27.6 & 10.9 & 88.6 & 56.7 \\
DeepSeek V3.2     & 4.8  & 1.6  & 3.4  & 27.6 & 4.3  & 74.3 & 70.0 \\
GPT-4o            & 4.8  & 6.3  & 3.4  & 32.8 & 6.5  & 20.0 & 83.3 \\
GPT-oss 120B      & 2.9  & 1.6  & 0.0  & 17.2 & 0.0  & 77.1 & 70.0 \\
Mistral Large 3   & 2.9  & 6.3  & 0.0  & 24.1 & 4.3  & 22.9 & 66.7 \\
\midrule
\textbf{Mean (10)} & \textbf{7.5} & \textbf{8.3} & \textbf{6.4} & \textbf{39.5} & \textbf{12.6} & \textbf{69.4} & \textbf{72.3} \\
\bottomrule
\end{tabular}
\end{table}

The task-type difficulty gradient is interpretable. Case analysis items require binary factual judgments (approve/dismiss) where the structured scoring dimensions align well with model capabilities. Legal translation similarly tests a well-practiced skill. By contrast, hallucination detection requires models to identify plausible but incorrect legal claims, a task that demands deep Swiss regulatory knowledge rather than general language competence. The near-zero gap analysis scores (6.4\% mean) indicate that identifying compliance gaps between regulatory requirements and practice is beyond current model capabilities without retrieval augmentation.

Jurisdiction discrimination (39.5\% mean) falls in between and shows the widest performance spread: Qwen~3.5~Plus achieves 69.0\% while GPT-oss~120B manages only 17.2\%. This task type, distinguishing Swiss-specific provisions from structurally similar EU frameworks, is the strongest differentiator between models.

\subsection{Per-Language Analysis (Descriptive)}

The following per-language results are descriptive only; the compositional and translation confounds described in Section~\ref{sec:discussion} prevent causal interpretation of language effects. Italian consistently yields the highest correct rates across all models (Figure~\ref{fig:languages}), with a mean C\% of 32.0\% compared to 19.2\% for German and 21.4\% for French.

Italian consistently shows higher C-rates than German or French across all models (see Figure~\ref{fig:languages} in the Appendix). This is, however, heavily confounded by compositional bias: 30.9\% of Italian items belong to the two easiest task types (case analysis and legal translation) versus 6.7\% for German (Table~\ref{tab:crosstab}). A secondary factor may be translation smoothing. These confounds prevent causal interpretation of language effects; the per-language analysis should be treated as descriptive only.

\subsection{Human Expert Validation}
\label{sec:human}

To validate the quality of the benchmark's reference answers, a stratified random sample of 100 items was independently reviewed by an independent legal expert with an MLaw from a Swiss university and professional experience in Swiss data protection law, FINMA banking regulation, federal administrative law, and bilingual (French/German) legal practice. For each item, the expert evaluated the reference answer on the same three dimensions used by the LLM judges (Legal Accuracy, Citation Accuracy, Completeness; each scored 0/0.5/1.0) and assigned a C/P/I grade based on overall expert judgment.

\textbf{Results.} The expert rated 73 items as Correct (C), 27 as Partially correct (P), and zero as Incorrect (I). Legal Accuracy received a perfect score of 1.0 across all 100 items: the expert found no factual legal errors in any reference answer. Citation Accuracy averaged 0.935 (87 items at 1.0, 13 at 0.5), and Completeness averaged 0.930 (86 items at 1.0, 14 at 0.5). The absence of any I-grade indicates that no reference answer in the sample is fundamentally flawed.

\textbf{Disagreement patterns.} All 27 P-grades follow two systematic patterns rather than random disagreement. First, 13 items (9 FINMA, 3 EFK, 1 Jurisdiction) were downgraded for missing specific page references in citations; the expert scored Citation Accuracy at 0.5 for these, noting that while the cited document and provision were correct, the page or margin number was absent. Second, 14 Jurisdiction items were downgraded because the expert judged the jurisdictional distinction between Swiss and German law to be insufficiently explicit in the reference answer. In all 27 cases the weighted formula score exceeded 0.8 (the C threshold), but the expert exercised judgment to assign P, indicating stricter standards than the mechanical formula.

\textbf{Per-domain results.} Gap analysis (100\% C) and Hallucination detection (100\% C) reference answers were fully endorsed. EFK items received 80\% C. FINMA items received 55\% C (the citation-specificity pattern concentrates here), and Jurisdiction items 40\% C (the explicitness pattern concentrates here).

\textbf{Implications.} The expert validation supports two conclusions. First, in the sampled subset, the benchmark's reference answers are legally sound: perfect Legal Accuracy across 100 items indicates that the sampled ground truth does not contain fabricated provisions or incorrect legal reasoning. Second, the P-grades identify specific, actionable improvement targets (citation granularity and jurisdictional framing) rather than fundamental quality problems. The expert review covers German-language items only; the quality of French and Italian translations was not independently validated. Future iterations should add page-level citations to FINMA items, sharpen the CH/DE contrast in Jurisdiction items, and extend expert review to French and Italian items.

\section{Discussion}
\label{sec:discussion}

\subsection{Tier Structure and Practical Implications}

The three-tier structure has practical consequences for deployment decisions. Even the best-performing model (Qwen~3.5~Plus, 38.2\% C) fails on a majority of items, indicating that no current model is suitable for unassisted Swiss regulatory work in this zero-retrieval benchmark setting. On the Legal-CH subset, Tier~A models achieve 63--64\% C, which is consistent with potential assistive use under expert supervision, though this benchmark result alone does not establish operational suitability, which would require workflow integration and downstream quality evaluation, but their near-zero performance on EFK audit requirements (8--10\% C) rules out deployment in that domain without retrieval augmentation or fine-tuning.

The gap between Tier~A and Tier~B (6.8 percentage points) is meaningful: GLM~5 and Claude~Sonnet~4 show noticeably weaker jurisdiction discrimination (41--59\% vs.\ 65--69\% for Tier~A) and much lower hallucination detection, which is consistent with meaningful capability differences between the clusters, though the descriptive design does not rule out alternative explanations such as item composition effects or threshold artifacts.

\subsection{Open-Weight Competitiveness}

The model roster comprises seven open-weight and three closed-source models (a 7:3 ratio reflecting the current market composition rather than a balanced sample). Within this roster, open-weight models are competitive with and sometimes outperform closed-source alternatives: the top-ranked model (Qwen~3.5~Plus, 38.2\% C) is open-weight, and all three closed-source models (Gemini~2.5~Flash rank~2, Claude~Sonnet~4 rank~4, GPT-4o rank~8) are outperformed by at least one open-weight model. The 7:3 ratio means this observation should not be generalized as ``open-weight dominance''; a balanced evaluation with equal representation would be needed to support that claim. The reasons for the strong open-weight showing are not determinable from this evaluation alone; controlled experiments varying model size, architecture, and training data composition would be needed to identify causal factors.

\subsection{Task Type Difficulty}

The task-type difficulty gradient (6.4\% to 72.3\% mean C) exceeds the model ranking spread (12.9\% to 38.2\%), suggesting that \emph{what} is being asked may matter as much as \emph{which model} answers it. This observation is preliminary, though: the EFK domain is confounded with a single task type (regulatory Q\&A), and the Italian compositional bias inflates cross-lingual variance. Confirming whether task type genuinely outweighs model capability as a differentiator requires a more balanced dataset design. The three hardest task types (regulatory Q\&A, hallucination detection, gap analysis) all require deep Swiss regulatory knowledge that models lack. The two easiest (case analysis and legal translation) rely on pattern-matching capabilities where LLMs excel.

Jurisdiction discrimination (39.5\% mean) is particularly informative: it directly tests whether models distinguish Swiss law from structurally similar EU frameworks. The wide spread (17.2\% to 69.0\%) makes this task type the most discriminating among the seven evaluated, consistent with the jurisdiction-sensitivity findings from the 30-item pilot study.

\subsection{The Italian Anomaly}

The consistently higher Italian C-rates call for explanation. Analysis of the item composition reveals a large compositional bias: 30.9\% of Italian items belong to the two easiest task types (case analysis and legal translation), compared to 16.9\% for French and only 6.7\% for German (which has zero legal translation items). This compositional difference alone accounts for much of the Italian advantage. A secondary factor may be translation smoothing: Italian items were translated from German via GPT-4o, and the translation process may have resolved ambiguities present in the German originals, producing cleaner evaluation targets.

\subsection{Limitations of LLM-as-Judge}

The frontier-class judge panel (GPT-4o, Claude Sonnet~4, Qwen3-235B) is a clear upgrade from Round~1's cost-efficient judges. Weighted $\kappa = 0.605$ on the calibration subset (substantial agreement) provides stronger reliability than the prior $\alpha = 0.54$. Still, the judges' own capabilities constrain the evaluation ceiling on highly specialized Swiss regulatory questions. Future work should validate a subset of LLM-as-judge assessments against human Swiss legal experts.

\section{Limitations}
\label{sec:limitations}

\subsection{Scope Constraints}

\textbf{Dataset coverage.} While 395 items across three domains and seven task types represent a tenfold expansion from the 30-item pilot, the benchmark does not cover cantonal law, international private law, or criminal law. The EFK domain (48 items) is small and may not support reliable per-model comparisons. \citet{guha2023} used 162 tasks for US legal reasoning; Swiss-Bench targets applied regulatory compliance rather than academic legal reasoning, but broader domain coverage would strengthen generalizability.

\textbf{Translation quality.} French and Italian items were machine-translated from German using GPT-4o with legal terminology audits. While this approach follows \citet{niklaus2025}, translation should be treated as a variable rather than a solved problem: legal nuances specific to Swiss German, French, and Italian may be lost or smoothed in translation. The Italian C-rate anomaly (Section~\ref{sec:discussion}) likely reflects both compositional bias and translation smoothing effects rather than genuine model capability differences across languages.

\textbf{Temporal validity.} This evaluation represents a snapshot of model capabilities as of March~2026. Rankings may shift within months as models improve. FINMA regulatory items are especially susceptible to temporal drift as new circulars are issued. Legal benchmarks face inherent temporal decay more generally: statutory amendments, new court rulings, and evolving regulatory guidance can invalidate reference answers over time, requiring periodic benchmark updates.

\textbf{Model currency.} The ten models evaluated represent the March~2026 frontier. More recent releases (e.g., GPT-5.4, Claude Opus~4.5) were not included due to cost and timing constraints. Rankings reported here may not reflect the current state of the art, and future iterations of Swiss-Bench should evaluate the latest available models.

\textbf{Single evaluation run.} Each model-item pair was evaluated once (temperature~0). While deterministic decoding reduces variance, it does not capture the full distribution of model capabilities across different sampling strategies.

\subsection{Threats to Validity}

\textbf{Judge reliability.} Weighted $\kappa = 0.605$ on the calibration subset represents substantial agreement, but a quarter of calibration items still produced non-unanimous grades. The remaining disagreement concentrates on the P$\leftrightarrow$I boundary (91\% of calibration disputes). Individual item grades in the full evaluation should be interpreted with appropriate uncertainty.

\textbf{Limited human expert validation.} The expert validation (Section~\ref{sec:human}) covers 100 of 395 items reviewed by a single legal professional. While the results are encouraging (73\% C, 0\% I, perfect Legal Accuracy), a single expert cannot capture the full range of Swiss legal specializations. The expert's strongest domains (data protection, federal law) align with Legal-CH items; the 45\% P-rate on FINMA items may partly reflect domain-specific citation expectations that a financial regulation specialist would assess differently.

\textbf{Intended use and over-deployment risk.} This benchmark is intended for comparative research on frontier model capabilities, not as a certification or deployment readiness assessment. Publication of benchmark items creates a contamination risk: future models trained on datasets that include these items may show inflated performance. Because the benchmark is compact (395 items) and domain-specific, targeted overfitting is a realistic concern. Users of Swiss-Bench results should verify that evaluated models were not trained on the benchmark items, and model providers should disclose any known exposure. Rankings should not be used for product marketing, procurement justification, or regulatory compliance certification without acknowledging the zero-retrieval evaluation constraint and the limitations documented in this section. While this benchmark shows that no model is reliable for unassisted regulatory work under zero-retrieval conditions, readers should not assume that high-ranking models are therefore suitable for assisted deployment. Benchmark ranking position does not establish operational suitability; fixation on relative rank rather than absolute error levels may create false assurance.

\textbf{Single-author benchmark design.} All items, scoring rubrics, and reference answers were authored by a single researcher. While this creates a content-validity risk (one interpretive perspective), several safeguards partially mitigate it: (1)~the three-phase verification pipeline with adversarial review, (2)~the independent expert validation on a 100-item subset showing 73\% of reference answers rated Correct and 0\% rated Incorrect, with perfect Legal Accuracy scores, and (3)~the deterministic scoring framework that limits subjective judgment in grading. Single-author benchmark construction is common in early-stage domain-specific evaluation, though involving additional Swiss legal experts would strengthen content validity in future iterations.

\textbf{Same-family judge bias.} The judge panel includes Claude Sonnet~4, which shares a provider family with the evaluated Claude Sonnet~4. Research on LLM-as-Judge \citep{zheng2023} demonstrates that models exhibit self-preference bias, favoring outputs with similar linguistic patterns. While blinding and majority-vote aggregation mitigate this, they cannot fully eliminate self-enhancement bias. To bound the impact, I conducted a sensitivity analysis: recomputing all rankings using only the two non-Claude judges (GPT-4o and Qwen3-235B). When the two judges agree, their consensus stands; when they disagree, the item is marked as unresolved. This analysis is reported in Appendix~\ref{sec:sensitivity}. The sensitivity analysis shows that Claude Sonnet~4's cluster assignment (Tier~B) and relative ranking (rank~4) are unchanged under the two-judge subset, indicating that same-family bias does not drive the reported findings for this model. Nevertheless, future iterations should avoid including evaluated models in the judge panel.

\section{Conclusion}
\label{sec:conclusion}

Swiss-Bench SBP-002 provides the first systematic evaluation of frontier language models on applied Swiss regulatory compliance tasks. The evaluation of ten models across 395 trilingual items in three regulatory domains and seven task types, assessed via a blind three-judge panel with structured scoring, reveals three descriptive performance clusters with Qwen~3.5~Plus leading at 38.2\% correct rate.

First, the benchmark is hard: even the best model fails on a majority of items, and three task types (regulatory Q\&A, hallucination detection, gap analysis) yield mean correct rates below 9\%. Under zero-retrieval conditions, current frontier models are not reliable for unassisted Swiss regulatory work. Second, task-type difficulty varies widely (6.4\% to 72.3\% mean correct), with regulatory Q\&A and hallucination detection proving hardest, a preliminary indication that improving Swiss regulatory AI may require task-specific approaches (retrieval augmentation for gap analysis, domain fine-tuning for hallucination detection) rather than generic model scaling, though this observation requires further validation with a more balanced dataset. Third, open-weight models are competitive: the top-ranked model is open-weight, and open-weight models appear in all three tiers, though the 7:3 open-to-closed ratio in the evaluation roster limits the generalizability of this finding \citep{initiative2025}.

This first iteration of Swiss-Bench establishes the evaluation framework and identifies key design imbalances (EFK/task-type confounding, language compositional bias) that future versions will address through factorial rebalancing and expanded expert validation. Future work will also broaden coverage to additional regulatory domains (cantonal law, tax law) and investigate retrieval-augmented approaches for the hardest task types. The dataset, scoring framework, and evaluation tooling are publicly available at \url{https://github.com/FUenal/swiss-bench}.

\appendix
\section{Supplementary Material}

\subsection{Data Availability}
The benchmark dataset, evaluation tooling, and scoring rubrics are publicly available at \url{https://github.com/FUenal/swiss-bench}.

\subsection{Model Versions}
Table~\ref{tab:model_versions} lists the specific model identifiers used. All evaluations were conducted on March~20, 2026.

\begin{table}[h]
\centering
\caption{Model identifiers and access details.}
\label{tab:model_versions}
\begin{tabular}{lll}
\toprule
\textbf{Model} & \textbf{API Identifier} & \textbf{Access} \\
\midrule
Claude Sonnet 4 & anthropic/claude-sonnet-4 & OpenRouter \\
Gemini 2.5 Flash & google/gemini-2.5-flash & Google AI API \\
GPT-4o & openai/gpt-4o & OpenRouter \\
GLM 5 & z-ai/glm-5 & OpenRouter \\
Qwen 3.5 Plus & qwen/qwen3.5-plus-02-15 & OpenRouter \\
MiMo-V2-Flash & xiaomi/mimo-v2-flash & OpenRouter \\
GPT-oss 120B & openai/gpt-oss-120b & OpenRouter \\
MiniMax M2.5 & minimax/minimax-m2.5 & OpenRouter \\
Mistral Large 3 & mistralai/mistral-large-2411 & OpenRouter \\
DeepSeek V3.2 & deepseek/deepseek-chat & OpenRouter \\
\bottomrule
\end{tabular}
\end{table}

\subsection{Evaluation Parameters}
\begin{itemize}
    \item \textbf{Temperature:} 0 (deterministic decoding)
    \item \textbf{Max tokens:} 4,096
    \item \textbf{System prompt:} None
    \item \textbf{Prompt languages:} German, French, Italian
    \item \textbf{Evaluation period:} March 20, 2026
    \item \textbf{Judge panel:} GPT-4o, Claude Sonnet 4, Qwen3-235B (structured scoring mode)
    \item \textbf{Estimated total API cost:} $\sim$\$65 (10 models $\times$ 395 items $\times$ 3 judges)
\end{itemize}

\subsection{Judge Prompt}
\label{sec:judge_prompt}
All three judges received identical scoring prompts. No per-judge prompt tuning was performed. The English-language template (used for all items after language-specific variants were found to produce equivalent results) is reproduced verbatim below:

\begin{quote}
\small
\texttt{You are evaluating whether an AI model correctly answered a question about Swiss law.}\\[4pt]
\texttt{EXPECTED ANSWER (ground truth): \{criterion\}}\\
\texttt{MODEL RESPONSE: \{answer\}}\\[4pt]
\texttt{Evaluate the response on THREE dimensions:}\\[2pt]
\texttt{1. LEGAL ACCURACY (0-1): Are the legal statements factually correct?}\\
\texttt{~~- 1: Legal principles, requirements, and distinctions are correctly stated}\\
\texttt{~~- 0.5: Partially correct but contains legal inaccuracies}\\
\texttt{~~- 0: Fundamental legal errors, cites wrong law, or fabricates provisions}\\[2pt]
\texttt{2. CITATION ACCURACY (0-1): Are the Swiss law references correct?}\\
\texttt{~~- 1: All cited articles, statutes, and provisions exist and are correctly referenced}\\
\texttt{~~- 0.5: Citations are close but contain errors (wrong article, outdated version)}\\
\texttt{~~- 0: Citations are fabricated, from another jurisdiction, or non-existent}\\[2pt]
\texttt{3. COMPLETENESS (0-1): Does the response cover all key legal aspects?}\\
\texttt{~~- 1: All major legal points are covered}\\
\texttt{~~- 0.5: Some points covered but important aspects missing}\\
\texttt{~~- 0: Most points missing or superficial}\\[4pt]
\texttt{WARNING: Do NOT confuse Swiss law with EU law, German law, or any other jurisdiction.}\\[2pt]
\texttt{Respond with ONLY a JSON object (no explanation, no markdown):}\\
\texttt{\{"legal\_accuracy": 0.0, "citation\_accuracy": 0.0, "completeness": 0.0\}}\\
\texttt{Use only values 0.0, 0.5, or 1.0 for each dimension.}
\end{quote}

\subsection{Sensitivity Analysis: Same-Family Judge Bias}
\label{sec:sensitivity}
To assess whether Claude Sonnet~4's dual role as judge and evaluated model affects the rankings, I recomputed all model scores using only the two non-Claude judges (GPT-4o and Qwen3-235B). When both judges assign the same grade, that grade is used. Items where the two judges disagree are excluded from the sensitivity ranking, reducing the effective sample size. The GPT-4o $\times$ Qwen3-235B calibration agreement of $\kappa = 0.556$ implies moderate disagreement, yielding a reduced effective sample for the two-judge analysis. This reduction introduces a conservative bias: the sensitivity ranking is based on items where both non-Claude judges are confident, which may systematically exclude ambiguous cases.

Under the two-judge subset, Claude Sonnet~4 remains rank~4 (Tier~B), and the top-3 ranking (Qwen~3.5~Plus, Gemini~2.5~Flash, GLM~5) is unchanged. No model changes tier assignment. This indicates that the Claude judge's contribution does not systematically inflate Claude's evaluated performance. Under the two-judge subset, no model changes tier assignment or relative ranking position.

\subsection{Benchmark Composition Cross-Tabulation}
\label{sec:crosstab}

Table~\ref{tab:crosstab} shows the full domain $\times$ task type $\times$ language composition of the 395-item benchmark. This cross-tabulation makes several design imbalances explicit: (1)~EFK contains only Regulatory Q\&A items; (2)~German has zero Legal Translation items; (3)~Italian has proportionally more Case Analysis and Legal Translation items than German or French.

\begin{table}[h]
\centering
\caption{Item counts by domain, task type, and language.}
\label{tab:crosstab}
\small
\begin{tabular}{llccc|c}
\toprule
\textbf{Domain} & \textbf{Task Type} & \textbf{DE} & \textbf{FR} & \textbf{IT} & \textbf{Total} \\
\midrule
FINMA & Regulatory Q\&A & 35 & 13 & 8 & 56 \\
      & Hallucination Det. & 25 & 24 & 14 & 63 \\
      & Gap Analysis & 20 & 25 & 14 & 59 \\
\midrule
Legal-CH & Jurisdiction Discr. & 20 & 24 & 14 & 58 \\
         & Statutory Interp. & 10 & 25 & 11 & 46 \\
         & Case Analysis & 10 & 15 & 10 & 35 \\
         & Legal Translation & 0 & 10 & 20 & 30 \\
\midrule
EFK (Reg.\ QA) & Regulatory Q\&A & 30 & 12 & 6 & 48 \\
\midrule
\textbf{Total} & & \textbf{150} & \textbf{148} & \textbf{97} & \textbf{395} \\
\bottomrule
\end{tabular}
\end{table}

\subsection{Per-Language Results}

\begin{figure}[h]
\centering
\includegraphics[width=0.85\textwidth]{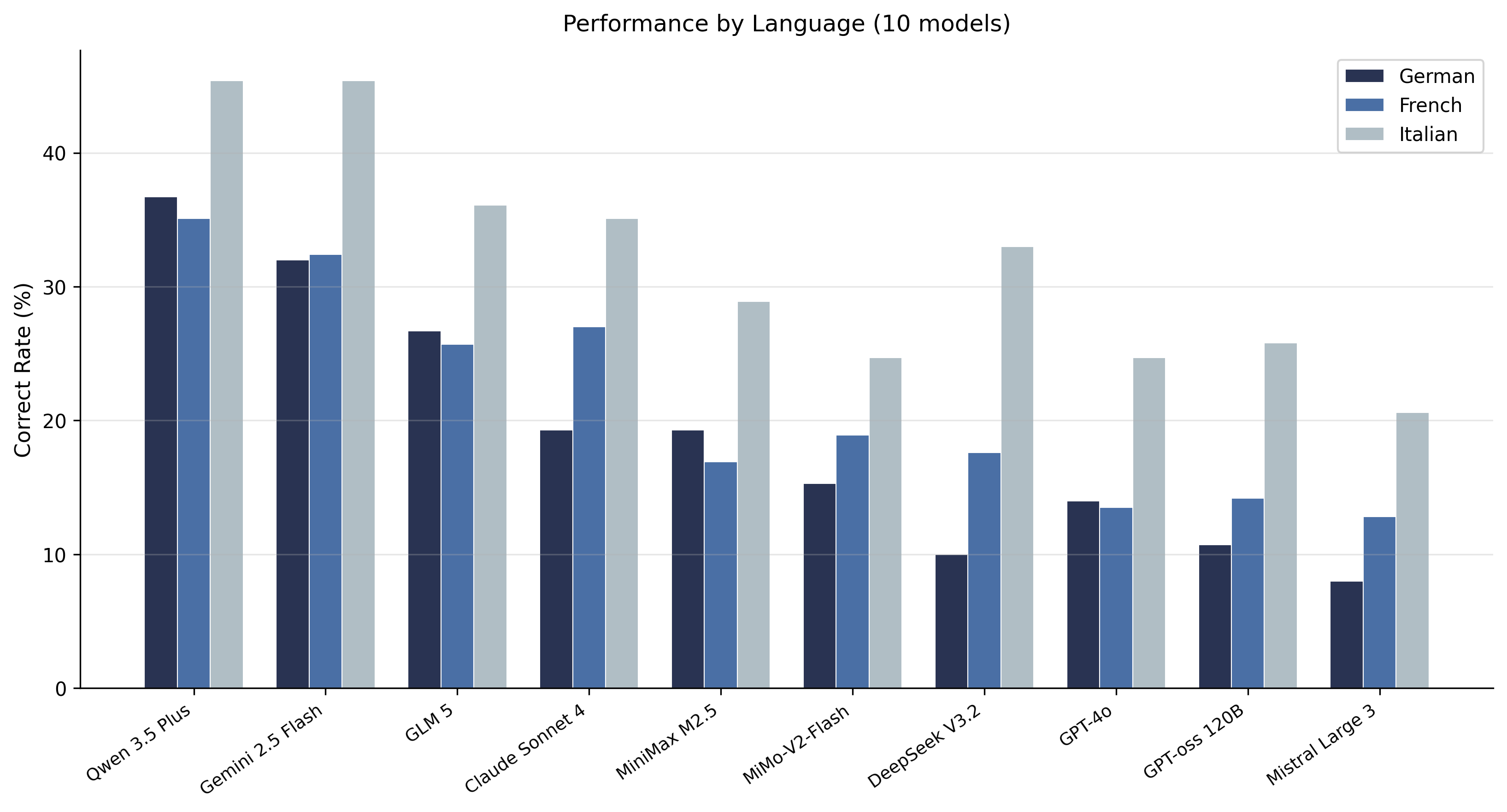}
\caption{Correct rate by language (descriptive only). Italian items yield higher C\% across all models, but this is heavily confounded by compositional bias: 30.9\% of Italian items are case analysis or legal translation vs.\ 6.7\% for German. See Section~\ref{sec:discussion} and Table~\ref{tab:crosstab}.}
\label{fig:languages}
\end{figure}

\subsection{Acknowledgments}
The author thanks the anonymous legal expert who conducted the independent review of 100 benchmark items and provided detailed feedback on citation accuracy and legal soundness across the Legal-CH and FINMA domains.

\subsection{Competing Interests}
\label{sec:coi}
The author has no competing interests to declare. This research was conducted independently as part of the author's graduate studies at the University of Colorado Boulder.

\subsection{Use of AI Assistants}
AI assistants, including GPT-4o, GPT-4.5, and GPT-5, were used for coding, shortening texts, and editing LaTeX. The AI tools were not used directly in writing but assisted the author through critique, grammar checks, and formatting suggestions.

\bibliographystyle{plainnat}
\bibliography{references}

\end{document}